\documentclass[11pt,a4paper]{article}

\usepackage{times}
\usepackage{latexsym}

\usepackage[bottom,hang]{footmisc}
\usepackage{microtype}

\usepackage{amsmath,amsfonts,bm}

\def\eqref#1{equation~\ref{#1}}

\def\1{\bm{1}}

\DeclareMathAlphabet{\mathsfit}{\encodingdefault}{\sfdefault}{m}{sl}
\SetMathAlphabet{\mathsfit}{bold}{\encodingdefault}{\sfdefault}{bx}{n}

\usepackage{mathabx}
\usepackage{times,latexsym}
\usepackage{url}
\usepackage[T1]{fontenc}
\usepackage{enumitem}
\usepackage{fancybox}
\usepackage{calc}
\usepackage{amsmath}
\usepackage{stmaryrd}
\usepackage{graphicx}       %
\usepackage{babel}
\usepackage{color}
\usepackage{multirow}
\usepackage{todonotes}
\usepackage{soul}
\usepackage{tabularx}
\usepackage{paralist} %
\usepackage{comment}

\newcommand{\remove}[1]{}
\newcommand{\pegpred}[1]{\texttt{#1}} 
\newcommand{\pegarg}[1]{\textit{#1}}
\newcommand{\peglabel}[1]{``#1''}
\newcommand{\ap}[0]{Autoprotocol}

\newcommand\tab[1][0.7cm]{\hspace*{#1}}
\newcommand{\affa}{{$^{\dagger}$}}
\newcommand{\affc}{{$^{\star}$}}
\newcommand{\affd}{{$^{\ddagger}$}}

\newcommand{\newWLP}{{X-WLP}\xspace} %
\newcommand{\corpusname}[0]{\newWLP}
\newcommand{\numannoprotos}{{279}\xspace} %
\newcommand{\numtokens}{{54.1K}\xspace} %
\newcommand{\numsents}{{3,708}\xspace} %

\newcommand{\numdoubleprotos}{{44}\xspace} %
\newcommand{\repourl}{\url{https://textlabs.github.io/}} %

\newcommand{\matsynth}{{MSPTC}\xspace}
\newcommand{\chemsynth}{{CSP}\xspace}

\usepackage[linesnumbered,ruled,vlined]{algorithm2e}

\usepackage{booktabs}

\SetCommentSty{mycommfont}

\SetKwInput{KwInput}{Input}                %
\SetKwInput{KwOutput}{Output}              %

\usepackage[hyperref]{eacl2021}

\aclfinalcopy %
\def\aclpaperid{875} %

\title{Process-Level Representation of Scientific Protocols \\with Interactive Annotation}

\author{Ronen Tamari\affa\thanks{Work begun on an internship at the Allen Institute for Artificial Intelligence.} \tab Fan Bai\affd \tab Alan Ritter\affd \tab {\bf Gabriel Stanovsky}\affa\affc  \\
  \affa The Hebrew University of Jerusalem \\
  \affd Georgia Institute of Technology \\
  \affc Allen Institute for Artificial Intelligence \\
  \texttt{\{ronent,gabis\}@cs.huji.ac.il} \\
  \texttt{\{fan.bai,alan.ritter\}@cc.gatech.edu} 
}

\date{}

\begin{document}
\maketitle
\begin{abstract}
We develop Process Execution Graphs~(PEG), a document-level representation of real-world wet lab biochemistry protocols, addressing challenges such as cross-sentence relations, long-range coreference, grounding, and implicit arguments. 
We manually annotate PEGs in a corpus of complex lab protocols  with  a novel interactive textual simulator that keeps track of entity traits and semantic constraints during annotation.
We use this data to develop graph-prediction models, finding them to be good at entity identification and local relation extraction, while our corpus facilitates further exploration of challenging long-range relations.\footnote{Our annotated corpus, simulator, annotation interface, interaction data, and models are available for use by the research community at \repourl{}.}

\end{abstract}

\section{Introduction}
\label{sec:intro}
There is a drive in recent years towards automating  wet lab environments, where menial benchwork procedures such as pipetting, centrifuging, or incubation are software-controlled, and either executed by fully automatic lab equipment~\cite{lee2018autoprotocol}, or with a human-in-the-loop~\citep{ben_keller_2019_2583232}.
These environments allow reliable and precise experiment reproducbility while relieving researchers from tedious and laborious work which is prone to human error~\cite{bates2017wet,prabhu2017dawn}.
To achieve this, several programmatic formalisms are developed to describe an experiment as an executable program.
For example, Autoprotocol~\cite{lee2018autoprotocol} defines a \pegpred{mix} predicate  taking three arguments: \pegarg{mode}, \pegarg{speed}, and \pegarg{duration}.

A promising direction to leverage automatic wet-lab environments is a conversion from 
natural language protocols, written in expressive
free-form language, to low-level instructions, ensuring a non-ambiguous,  repeatable description of experiments.

\begin{figure}[t!]
\centering
\includegraphics[width=\columnwidth]{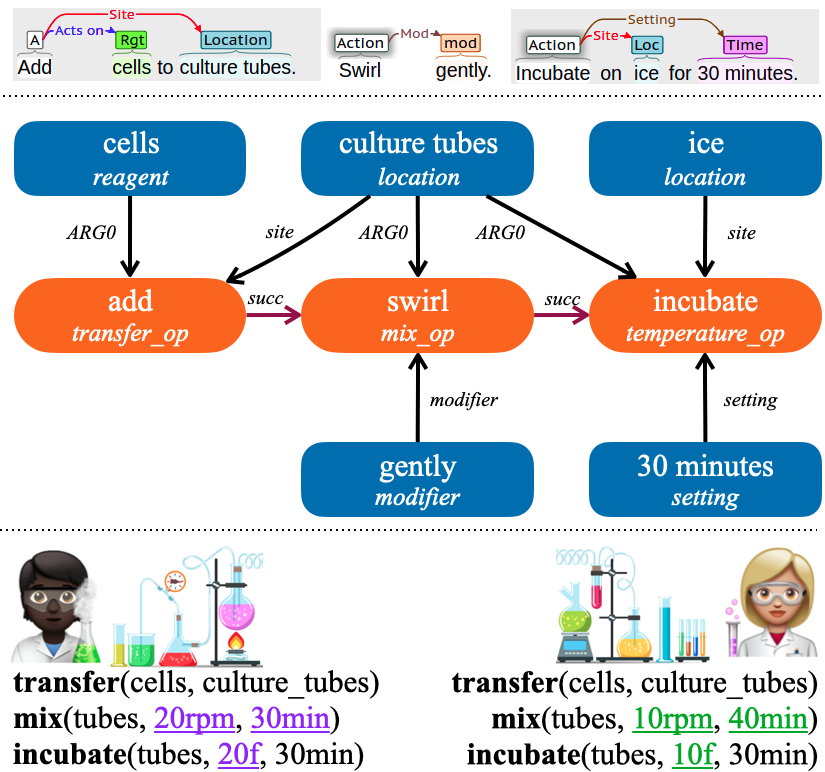}

\caption{\label{fig:example} 
We develop a scaffold (center) between sentence-level lab procedure representations (top) and low-level, lab-specific instructions (bottom). 
The Process Execution Graph (PEG) captures document-level relations between procedures (orange rounded nodes) 
and their arguments (blue rectangular nodes).
}
\end{figure}

In this work, we focus on a crucial first step towards such conversion -- the extraction and representation of the relations conveyed by the protocol in a formal graph structure, termed Process Execution Graphs (PEG), exemplified in  Figure~\ref{fig:example}.
PEGs capture both concrete, exact quantities (``\pegarg{30 minutes}''), as well as vague instructions (``swirl \pegarg{gently}'').
A researcher can then port the PEG (either manually or automatically) to their specific lab equipment, e.g., specifying what constitutes a gentle swirl setting and adding missing arguments, such as the temperature of the incubation in Figure~\ref{fig:example}.

Formally,  PEGs are directed, acyclic labeled graphs, capturing how objects in the lab (e.g., \pegarg{cells}, \pegarg{tubes}) are manipulated by lab operations (e.g., \pegpred{mixing}, \pegpred{incubating}), and in what order. Importantly, PEGs capture relations which may span across multiple sentences and implicit arguments. 
For example, the PEG in Figure~\ref{fig:example} explicitly captures the relation between \pegarg{culture tubes}, mentioned in the first sentence, and \pegpred{swirl} and \pegpred{incubate} which appear in later sentences.

\begin{figure}[t!]
\centering
\includegraphics[width=0.80\linewidth]{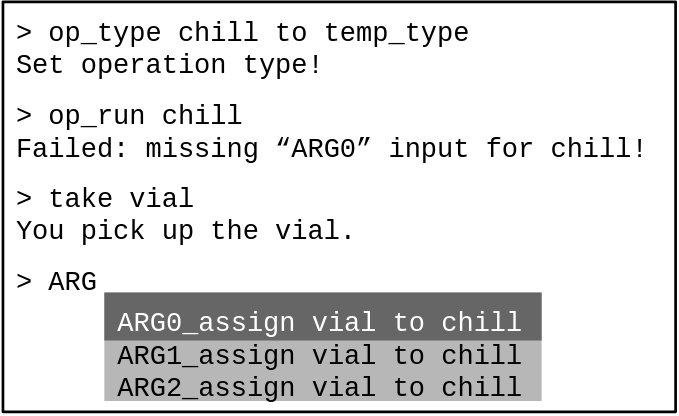}
\caption{\label{fig:interaction} Example interaction with our simulator, showing predicate grounding (``chill'' is a \pegpred{temp\_type} operation) input assignment (``vial'' is an argument of ``chill''), validation (warning for a missing argument) and auto-complete driven by state-tracking, where only legal instructions in a given state are presented.}
\end{figure}

To annotate long and complex lab protocols, we develop a text-based game annotation interface simulating objects and actions in a lab environment~(see example in Figure \ref{fig:interaction}).
Our annotators are given wet-lab protocols written in natural language taken from biochemistry publications, and are asked to repeat their steps by issuing textual commands to the simulator. The commands are deterministically converted to our PEG representation.
This interface takes much of the burden off annotators by keeping track of object traits and commonsense constraints.
For example, when the annotator issues a \pegpred{transfer} command for a container, the simulator moves all its contents as well.
We find that in-house annotators were able to effectively use this interface on complex protocols, achieving good agreement.

Finally, we use this data to explore several models, building upon recent advances in graph prediction algorithms~\cite{luan-etal-2019-general,wadden-etal-2019-entity}. We thoroughly analyze model performance and find that our data introduces interesting new challenges, such as complex co-reference resolution and long-range, cross-sentence relation identification.

In conclusion, we make the following contributions:
\begin{compactitem}
    \item We formalize a PEG representation for free-form, natural language lab protocols, providing a semantic scaffold between free-form scientific literature and low-level instruments instruction.
    \item We develop a novel annotation interface for procedural text annotation using text-based games, and show that it is intuitive enough for wet-lab protocol annotation by non-experts.
    \item We release \corpusname{}, a challenging corpus of \numannoprotos{} PEGs representing  document-level lab protocols. This size is on par with similar corpora of procedural text~\citep{dalvi-etal-2018-tracking,mysore2019materials,Vaucher2020}.
    \item We develop two graph parsers: a  pipeline model which chains predictions for graph sub-components, and a joint-model of mention and relation detectors.
\end{compactitem}

\section{Background and Motivation}
Several formalisms for programmatic lab controller interfaces were developed in recent years~\cite{yachie2017robotic,lee2018autoprotocol}.
For instance, \ap{} defines 35 lab commands, including \pegpred{spin}, \pegpred{incubate}, and \pegpred{mix}.\footnote{\url{https://autoprotocol.org/specification}}
While these define wet-lab experiments in a precise and unambiguous manner, they do not readily replace their natural language description in scientific publications, much like a model implementation in python does not replace its high-level description in ML papers. Similarly to ML model descriptions, lab protocols are often not specified enough to support direct conversion to low-level programs. For example, the protocol in Figure~\ref{fig:example} does not specify the swirling (mixing) speed or its duration.

Our process execution graph (PEG) captures the predicate-argument structure of the protocol, allowing it to be more lenient than a programming language (for example, capturing that \pegarg{gently} modifies \pegpred{swirl}). Better suited to represent underspecified natural language,
PEGs can serve as a convenient scaffold to support downstream tasks such as text-to-code assistants~\citep{Mehr101}. For example, by asking researchers to fill in missing required arguments  for \pegpred{swirl}.

To annotate PEGs,
we leverage the sentence-level annotations of~\newcite{kulkarni-etal-2018-annotated} (WLP henceforth). 
WLP, exemplified at the top of  Figure~\ref{fig:example}, collected sentence-level structures using the BRAT annotation tool~\cite{brat2012}. For example,  capturing that \pegarg{cells}, \pegarg{culture tubes} are arguments for \pegpred{add}. However, WLP does not capture cross-sentence implicit relations such that \pegarg{culture tubes} are an argument for \pegpred{incubate}. These are abundant in lab protocols, require tracking entities across many sentences, and are not easy to annotate using BRAT (see discussion in \S\ref{sec:data}).
We vastly extend upon WLP annotations, aiming to capture the full set of expressed protocol relations, using a novel text-based games annotation interface which lends itself to procedural text annotation.

\section{Task Definition: 
Process Execution Graphs}
\label{sec:task-def}
\begin{figure*}[t!]
\centering
\includegraphics[width=1\linewidth]{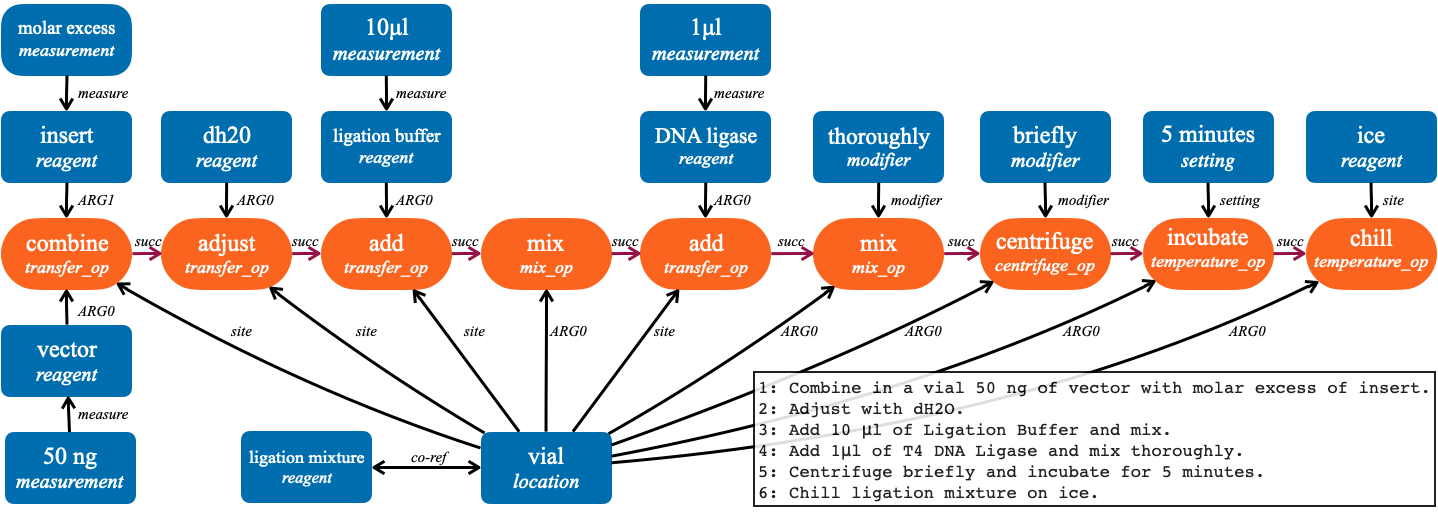}
\caption{\label{fig:full-process}
A full process gold PEG annotation from \corpusname{} for a real-world wet lab protocol whose text is presented in the lower right corner (protocol 512), exemplifying several common properties: 
(1) complex, technical language, in relatively short sentences;
(2) a chain of temporally-dependent, cross-sentence operations; 
(3) a common object that is being acted upon through side effects throughout the process (\pegarg{vial}); and
(4) \pegarg{vial} is mostly omitted in the text after being introduced in the first sentence, despite participating in all following sentences. In the last sentence it appears with a metonymic expression (\pegarg{ligation mixture}).
}
\end{figure*}

Intuitively, we extend the WLP annotations~\cite{kulkarni-etal-2018-annotated} from the sentence level to entire documents, aiming to capture \emph{all} of the relations in the protocol. Formally, our representation is a directed, labeled, acyclic graph structure, dubbed a Process Execution Graph (PEG), exemplified in Figures \ref{fig:example} and \ref{fig:full-process}, and formally defined below.

\begin{table}[b!]
\small
\setlength{\tabcolsep}{3pt}
\begin{tabular}{@{}llcc@{}}
\toprule
Operation type                                                   & Frequent example spans       & \multicolumn{1}{l}{Count} & \multicolumn{1}{l}{Pct.} \\ \midrule
Transfer                                                         & add, transfer, place         & 1301                      & 33.2                  \\
\begin{tabular}[c]{@{}l@{}}Temperature \\ Treatment\end{tabular} & incubate, store, thaw        & 503                       & 12.8                  \\
General                                                          & Initiate, run, do not vortex & 469                       & 11.9                  \\
Mix                                                              & mix, vortex, inverting       & 346                       & 8.8                   \\
Spin                                                             & spin, centrifuge, pellet     & 282                       & 7.2                   \\
Create                                                           & prepare, make, set up        & 178                       & 4.5                   \\
Destroy                                                          & discard, decant, pour off     & 170                       & 4.3                   \\
Remove                                                           & remove, elute, extract       & 168                       & 4.3                   \\
Measure                                                          & count, weigh, measure        & 149                       & 3.8                   \\
Wash                                                             & wash, rinse, clean           & 146                       & 3.7                   \\
Time                                                             & wait, sit, leave             & 114                       & 2.9                   \\
Seal                                                             & cover, seal, cap             & 68                        & 1.7                   \\
Convert                                                          & change, transform, changes   & 21                        & 0.5                   \\ \bottomrule
\end{tabular}
\caption{\label{tab:ops} Details of PEG predicate types, along with example frequent trigger spans and relative frequency in \corpusname{}.}
\end{table}

\paragraph{Nodes} PEG nodes are triggered by explicit text spans in the protocol, e.g., ``swirl", or ``ice''. Nodes consist of two types: 
(1) \emph{predicates}, marked in orange: denoting lab operations, such as \pegpred{add} or \pegpred{incubate}; and
(2) \emph{arguments}, marked in blue: representing physical lab objects (e.g., \pegarg{culture tubes}, \pegarg{cells}), exact quantities (\pegarg{30 minutes}), or abstract instructions (e.g., \pegarg{gently}).

\paragraph{Node grounding}
The PEG formulation above is motivated as a scaffold towards fully-executable lab programs employed in automatic lab environments. 
To achieve this, we introduce an ontology for each of the node types, based on the \ap{} specification~\cite{lee2018autoprotocol}, as indicated below each text span in Figures~\ref{fig:example} and \ref{fig:full-process}.
For example, \pegpred{swirl} corresponds to an \ap{} \pegpred{mix} operation, a \pegarg{culture tube} is of type \pegarg{location}, and \pegarg{30 minutes} is a \pegarg{setting}.
See Tables~\ref{tab:ops}, \ref{tab:ents} for details of predicate and argument types respectively, their frequencies in our data and example spans.

\begin{table}[]
\small
\setlength{\tabcolsep}{3pt}
\begin{tabular}{@{}llcc@{}}
\toprule
Argument type & Frequent example spans            & \multicolumn{1}{l}{Count} & \multicolumn{1}{l}{Pct.} \\ \midrule
Reagent     & supernatant, dna, sample          & 3362                      & 32.6                  \\
Measurement & 1.5 mL, 595nm, 1pmol              & 1924                      & 18.6                 \\
Setting     & overnight, room temperature       & 1622                      & 15.7                  \\
Location    & tube, ice, plates                 & 1373                      & 13.3                  \\
Modifier    & gently, carefully, clean     & 1070                      & 10.3                  \\
Device      & forceps, pipette tip  & 590                       & 5.7                   \\
Method      & dilutions, pipetting  & 271                       & 2.6                   \\
Seal        & lid, cap, aluminum foil           & 97                        & 0.9                   \\ \bottomrule
\end{tabular}
\caption{\label{tab:ents} Details of PEG argument types, along with example frequent trigger spans and relative frequency in \corpusname{}. }
\end{table}
\paragraph{Edges} 
Following PropBank notation~\cite{propbank}, PEGs consist of three types of edges derived from the \ap{} ontology, and  denoted by their labels:
(1) core-roles (e.g., \peglabel{ARG0}, \peglabel{ARG1}): indicating predicate-specific roles, aligning with \ap{}’s ontology. For example, \pegarg{ARG0} of \pegpred{mix} assigns the element to be mixed;
(2) non-core roles (e.g., \peglabel{setting}, \peglabel{site}, or \peglabel{co-ref}): indicate predicate-agnostic relations. For example, the \pegarg{site} argument always marks the location in which a predicate is taking place; and
(3) temporal edges, labeled with a special \peglabel{succ} label: define a temporal transitive ordering between predicates. In Figure~\ref{fig:example}, \pegpred{add} occurs before \pegpred{swirl}, which occurs before \pegpred{incubate}.
See Table~\ref{tab:core-rels} for predicate-specific core-role semantics, and Table~\ref{tab:graph-stats} for non-cores roles types and frequencies of all roles in \corpusname{}. See Appendix~\ref{ssec:roles-ontology} for the rules defining what relations can hold between various entity types.

\begin{table}[]
\small
\setlength{\tabcolsep}{3pt}
\begin{tabular}{@{}lll@{}}
\toprule
Operation                                                         & Role Semantics                                                                                                             & Required   \\ \midrule
Spin                                                             & \begin{tabular}[c]{@{}l@{}}ARG0 centrifuged to \\ produce solid phase\\ ARG1 and/or liquid \\ phase ARG2\end{tabular} & ARG0       \\
Convert                                                          & ARG0 converted to ARG1                                                                                                & ARG0, ARG1 \\
Seal                                                             & ARG0 sealed with ARG1                                                                                                 & ARG0       \\
Create                                                           & ARG* are created                                                                                                   & ARG0       \\
General                                                          & -                                                                                                                     & ARG0       \\
Destroy                                                          & ARG* discarded                                                                                                        & ARG0       \\
Measure                                                          & ARG* to be measured                                                                                                   & ARG0       \\
Mix                                                              & ARG* are mixed                                                                                                        & ARG0       \\
Remove                                                           & ARG0 removed from ARG1                                                                                                & ARG0       \\
\begin{tabular}[c]{@{}l@{}}Temperature \\ Treatment\end{tabular} & ARG* to be heated/cooled                                                                                              & ARG0       \\
Time                                                             & Wait after operation on ARG0                                                                                          & ARG0       \\
Transfer                                                         & \begin{tabular}[c]{@{}l@{}}ARG* are sources, \\ transferred to "site"\end{tabular}                                    & ARG0, site \\
Wash                                                             & ARG0 washed with ARG1                                                                                                 & ARG0 \\ \bottomrule
\end{tabular}
\caption{\label{tab:core-rels} Details of core role semantics for all operation types. The ``Required'' column specifies which roles must be filled for a given operation. ARG* is short for $\left\{ \text{ARG0}, \text{ARG1}, \text{ARG2}\right\}$. }
\end{table}

\paragraph{Relation to \ap{}}
As shown at the bottom of Figure~\ref{fig:example}, a PEG is readily convertible to \ap{} or similar laboratory interfaces  once it is fully instantiated, thanks to edge labels and node grounding to an ontology. For example, a researcher can specify what \pegarg{gently} means in terms of mixing speed for their particular lab instruments.

\paragraph{Reentrancies and cross-sentence relations} While the PEG does not form directed cycles,\footnote{This happens because the temporal relations define a partial ordering imposed by the linearity of the execution.} it does form non-directed cycles (or \emph{reentrancies}) -- where there exists nodes $u, v $ such that there are two different paths from $u$ to $v$. This occurs when an object participates in two or more temporally-dependent operations. For example, see \pegarg{culture tubes}, which participates in all operations in Figure~\ref{fig:example}. In addition, edges $(u, v)$ may be triggered either by \emph{within-sentence relations}, when both $u$ and $v$ are triggered by spans in the same sentence, or by \emph{cross-sentence relations}, when $u$ and $v$ are triggered by spans in different sentences. In the following section we will show that both reentrancies and cross-sentence relations, which are not captured by previous annotations, are abundant in our annotations.

\section{Data Collection: The \corpusname{} Corpus}
\label{sec:data}

\begin{table}[tb!]
\center
\resizebox{\columnwidth}{!}{
\begin{tabular}{@{}lcccc@{}}
\toprule
                     & \multicolumn{1}{l}{X-WLP (ours)} & \multicolumn{1}{l}{\matsynth} & \multicolumn{1}{l}{\chemsynth} & \multicolumn{1}{l}{ProPara} \\ \midrule
\# words             & 54k                       & 56k                    & 45k                      & 29k                         \\
\# words / sent.     & 14.6                      & 26                     & 25.8                      & 9                           \\
\# sentences         & 3,708                     & 2,113                  & 1,764                   & 3,300                        \\
\# sentences / docs. & 13.29                     & 9                      & N/A                      & 6.8                         \\
\# docs.             & 279                       & 230                    & N/A                   & 488                         \\ \bottomrule
\end{tabular}
}
\caption{\label{tab:stats} 
Statistics of our annotated corpus (\corpusname{}).
\corpusname{} annotates complex documents, constituting more than 13 sentences on average.
\corpusname{} overall size is on par with other recent procedural corpora, including
ProPara~\citep{dalvi-etal-2018-tracking}, material science~(\matsynth; \citet{mysore2019materials}) and chemical synthesis procedures~(\chemsynth;  \citet{Vaucher2020}). CSP is comprised of annotated sentences (document level information is not provided).} %
\end{table}

In this section, we describe in detail the creation of our annotated corpus:~\corpusname.
The protocols in \corpusname{} are a subset (44.8\%) of those annotated in the WLP corpus. These were chosen because they are covered well by \ap{}'s ontology (for details on ontology coverage, see \S\ref{ssec:ontology-construction}). 

In total, we collected \numsents sentences (\numtokens tokens) in \numannoprotos wet lab protocols annotated with our graph representation.
As can be seen in Table \ref{tab:stats}, \corpusname{} annotates long examples, often spanning dozens of sentences, and its size is comparable (e.g., in terms of annotated words) to the ProPara corpus~\citep{dalvi-etal-2018-tracking} and other related procedural datasets.

\subsection{WLP as a Starting Point}

Despite WLP's focus on sentence-level relations (see top of Figure~\ref{fig:example}), it is a valuable starting point for a document-level representation. We pre-populate our PEG representations with WLP's gold object mentions~(e.g., \pegarg{cells}, \pegarg{30 minutes}), operation mentions~(\pegpred{swirl} and \pegpred{incubate}), and within-sentence  relations~(e.g., between \pegarg{gently} and \pegpred{swirl}). We ask annotators to enrich them with type grounding for operations and arguments, as well as cross-sentence relations, as defined in \S\ref{sec:task-def}. From these annotations we obtain process-level representations as  presented in  Figures~\ref{fig:example} and \ref{fig:full-process}.

\subsection{Process-Level Annotation Interface: Text-Based Simulator}
\label{subsec:simulator}
Annotating cross-sentence relations and grounding without a dedicated user interface is an arduous and error-prone prospect. 
Consider as an example the \pegarg{ligation mixture} mention in Figure~\ref{fig:full-process}. This mention is a metonym for \pegarg{vial} (5 sentences earlier), after mixing in the \pegarg{ligase}. 
This kind of metonymic co-reference is known to be  difficult for annotation~\citep{Jurafsky+Martin:2009a}, 
and indeed, such complicated annotation has been a factor in the omission of cross-sentence information in similar domains~\citep{mysore2019materials}.
A simulator can provide a natural way to account for it by representing the relevant temporal and contextual information: after sentence 4, \pegarg{vial} contains the \pegarg{ligation buffer} mixed with other entities.

To overcome these challenges and achieve high-quality annotations for this complex task, we develop a simulator annotation interface, building upon the TextWorld framework~\citep{cote2018textworld}. 
This approach uses text-based games as the underlying simulator environment, which we adapt to the biochemistry domain.
The human annotator interacts with the text-based interface to simulate the raw wet lab protocol~(Figure~\ref{fig:interaction}): setting the types of operations (the first interaction sets the span ``chill" as a \pegpred{temperature} operation) and assigning their inputs (the last line assigns \pegpred{vial} as an input to \pegpred{chill}), while the simulator tracks entity states and ensures the correct number and type of arguments, based on the \ap{} ontology. For example, the second interaction in Figure~\ref{fig:interaction} indicates a missing argument for the \pegpred{chill} operation (the argument to be chilled). Finally, tracking temporal dependency (\peglabel{succ} edges) is also managed entirely by the simulator by tracking the order in which the annotator issues the different operations.

Further assistance is provided to annotators in the form of an auto-complete tool (last interaction in Figure~\ref{fig:interaction}), visualization of current PEG and a simple heuristic ``linter''~\cite{johnson1977lint} which flags errors such as ignored entities by producing a score based on the number of connected components in the output PEG.

See the project web page for the complete annotation guidelines, visualizations of annotated protocols, and demonstration videos of the annotation process.

\subsection{Data Analysis}
\label{subsec:iaa}

Four in-house CS undergraduate students with interest in NLP used our simulator to annotate the protocols of \corpusname{}, where \numdoubleprotos{} of the protocols were annotated by two different annotators to estimate agreement. 

\paragraph{Inter-annotator agreement.} We turn to the literature on abstract meaning representation~(AMR;~\citealp{Banarescu2013AbstractMR}) for established graph agreement metrics, which we adapt to our setting.
Similarly to our PEG representation, the AMR formalism has predicate and argument nodes (lab operations and entities in our notation) and directed labeled edges which can form undirected cycles through reentrancies~(nodes with multiple incoming edges).\footnote{Unfortunately, we cannot follow this analogy to train AMR models on our graphs, since, to the best of our knowledge, they are currently limited to single sentences, notwithstanding a promising recent initial exploration into multi-sentence AMR annotation~\cite{ogorman-etal-2018-amr}.}
In Table~\ref{tab:iaa} we report a graph Smatch score~\cite{cai-knight-2013-smatch} widely used to quantify AMR's graph structure agreement, as well as finer grained graph agreement metrics, adapted from~\citet{damonte-etal-2017-incremental}.
Smatch values are comparable to those obtained for AMR, where reported gold agreement varies between $0.69 - 0.89$~\cite{cai-knight-2013-smatch}, while our task deals with longer, paragraph length representations. Reentrancies are the hardest for annotators to agree on, probably since they involve longer-range, typically cross-sentence relations. On the other hand, local decisions such as argument and predicate identification achieve higher agreement, and also benefit greatly from the annotations of WLP.

\begin{table}[tb!]
\small
\center
\begin{tabular}{@{}lr@{}}
\toprule
\textbf{Agreement Metric}        & \textbf{F1}    \\ \midrule
Smatch                  & 84.99 \\
Argument identification   & 89.72 \\
Predicate identification & 86.68 \\
Core roles      & 80.52 \\
Re-entrancies            & 73.12 \\ \bottomrule
\end{tabular}

\caption{\label{tab:iaa} 
\corpusname{} inter-annotator agreement metrics. Smatch~\cite{cai-knight-2013-smatch} quantifies overall graph structure. Following metrics provide a finer-grained break down~\cite{damonte-etal-2017-incremental}.
} 
\end{table}
\begin{table}[tb!]
\small
\setlength\tabcolsep{3pt}
\center
\resizebox{\columnwidth}{!}{
\begin{tabular}{@{}lllll@{}}
\toprule
Relation           & \# Intra.    & \# Inter.   & Total & \# Re-entrancy \\ \midrule
Core               &              &             &       &                \\
{ $\bullet$ ARG0}               & 2962         & 952         & 3914  & 1645           \\
{ $\bullet$ ARG1 }              & 560          & 127         & 687   & 3              \\
{ $\bullet$ ARG2 }             & 84           & 123         & 207   & 77             \\
Total (core) & 3606   & 1202  & 4808  & 1725    \\ \midrule
Non-Core           &              &             &       &                \\
{ $\bullet$ site }              & 1306         & 325         & 1631  & 360            \\
{ $\bullet$ setting}            & 3499         & 2           & 3501  & -              \\
{ $\bullet$ usage }             & 1114         & 24          & 1138  & -              \\
{ $\bullet$ co-ref}            & 129          & 1575        & 1704  & -              \\
{ $\bullet$ located-at}        & 199          & 72          & 271   & -              \\
{ $\bullet$ measure }           & 2936         & 18          & 2954  & -              \\
{ $\bullet$ modifier}           & 1861         & 2           & 1863  & -              \\
{ $\bullet$ part-of}           & 72           & 65          & 137   & -              \\
Total (non-core)   & 11116        & 2083        & 13199 & 360            \\ \midrule
Temporal           & 1218                          & 788                           & 2006                      & -                                  \\ \midrule
Grand Total        & 15940 (80\%)                  & 4073 (20\%)                   & 20013                     & 2085                               \\ \bottomrule
\end{tabular}
}
\caption{\label{tab:graph-stats} Breakdown of PEG relation types by frequency in \corpusname{}, showing counts of inter/intra-sentence relations. Re-entrancies are possible only for core and \peglabel{site} arguments, and may be either inter or intra-sentence.}
\end{table}

\paragraph{Information gain from process-level annotation.}
Analysis of the relations in \corpusname{}, presented in Table~\ref{tab:graph-stats}, reveals that a significant proportion of arguments in PEGs are re-entrancies (32.4\%) or cross-sentence (50.3\%).\footnote{For these calculations we consider only argument relations that can in principle occur as re-entrancies: \peglabel{ARG*} and \peglabel{site}, see relation ontology in Appendix~\ref{ssec:roles-ontology} for details. Cross-sentence calculation includes co-reference closure information.}
Figure~\ref{fig:full-process} shows a representative example, with the \pegarg{vial} participating in multiple re-entrancies and long-range relations, triggered by each sentence in the protocol.
These relations are crucial to correctly model the protocols at the process level, and are inherently missed by sentence-level formalisms, showing the value of our annotations. 

To shed light on the additional process-level information captured by our approach relative to WLP, in Table \ref{tab:imp-stats} we compare the average number of arguments per operation node as well as the amount of operation nodes with no core arguments. For example, see the \pegpred{swirl} instruction at the top of Figure \ref{fig:example}: in WLP, this predicate has no core role argument and is thus semantically under-defined. \corpusname{} correctly captures the core role of \pegarg{culture tubes}.  By definition, our use of input validation by the simulator prevents semantic under-specification, which is likely a significant factor in the higher counts for cross-sentence relations and overall average arguments in \corpusname{}. 
\begin{table}[tb!]
\small
\setlength\tabcolsep{3pt}
\center
\resizebox{\columnwidth}{!}{
\begin{tabular}{@{}lcccc@{}}
\toprule
Dataset & \multicolumn{1}{l}{Avg. \#args/op} & \multicolumn{1}{l}{\#Ops. w/o core arg.} & \multicolumn{1}{l}{\#Ops.} & \multicolumn{1}{l}{Pct.} \\ \midrule
WLP     & 1.87                               & 3297                                       & 17485                          & 18.9                     \\
X-WLP   & 3.01                               & 0                                          & 3915                           & 0.0                      \\ \bottomrule
\end{tabular}
}
\caption{\label{tab:imp-stats} Comparison of average arguments per operation and percentage of semantically under-specified operations (missing core arguments) in WLP and \corpusname{}. }
\end{table}

\paragraph{Annotation cost.} 
The time to annotate an average document of 13.29 sentences was approximately 53 minutes (roughly 4 minutes per sentence), not including annotator training.
Our annotator pay was 13 USD / hour. The overall annotation budget for \corpusname{} was roughly 3,200 USD.

\section{Models}
\label{sec:models}
We present two approaches for PEG prediction.
First, in \S\ref{sec:pipeline} we design models for separate graph sub-component prediction, which are chained to form a pipeline PEG prediction model. 
Second, in \S\ref{sec:multitask} we present a model which directly predicts the entire PEG using a span-graph prediction approach.

\subsection{Pipeline Model (\textsc{Pipeline})}
\label{sec:pipeline}

A full PEG representation as defined in \S\ref{sec:task-def} can be obtained by chaining the following models which predict its sub-components. In all of these, we use SciBERT~\cite{beltagy-etal-2019-scibert} which was trained on scientific texts similar to our domain.

\paragraph{Mention identification.}
Given a scientific protocol written in natural language, we begin by identifying all experiment-involved text spans mentioning lab operations (predicates) or entities and their traits (arguments), which are the building blocks for PEGs. We model this problem of mention identification as a sequence tagging problem. Specifically, we transfer span-level mention labels, which are annotated in the WLP corpus into token-level labels using the BIO tagging scheme, then fine-tune the SciBERT model for token classification.

\paragraph{Predicate grounding.}
Next, we ground predicate nodes into the operation ontology types discussed in \S\ref{sec:task-def}. See Table~\ref{tab:ops} in the Appendix for the complete list.
Predicted mentions are marked using special start and end tokens (\texttt{[E-start]} and \texttt{[E-end]}), then fed as input to SciBERT.  The contextual embedding of \texttt{[E-start]} is input to a linear softmax layer to predict the fine-grained operation type.

\paragraph{Operation argument role labeling.}
Once the operation type is identified, we predict its semantic arguments and their roles.  Given an operation and an argument mention, four special tokens are used to specify the positions of their spans~\cite{baldini-soares-etal-2019-matching}.  Type information is also encoded into the tokens, for example, when the types of the operator and its argument are \texttt{mix-op} and \texttt{reagent} respectively, four special tokens \texttt{[E1-mix-op-start]}, \texttt{[E1-mix-op-end]}, \texttt{[E2-rg-start]} and \texttt{[E2-rg-end]} are used to denote the spans of the mention pair.  After feeding the input into SciBERT, the contextualized embeddings of \texttt{[E1-op-mix-start]} and \texttt{[E2-rg-start]} are concatenated as input to a linear layer that is used to predict the entity's argument role. %
Arguments of an operation can be selected from anywhere in the protocol, leading to many cross-sentence operation-argument link candidates. 
To accommodate cross-sentence argument roles, we use the entire document as input to SciBERT for each mention pair.  However, SciBERT is limited to processing sequences of at most 512 tokens.  To address this limitation, longer documents are truncated in a way that preserves surrounding context, when encoding mention pairs.\footnote{Given an input document, which has more than 512 words, with $n$ words between two mentions, we truncate the context to keep at most $(512 - n)/2$ words for each side.} Only 8 of the 279 protocols in our dataset contain more than 512 tokens.

\paragraph{Temporal ordering.}
Finally, we model order of operations using the \texttt{succ} relation~(see Figure~\ref{fig:full-process}).
These are predicted using a similar approach as argument role labeling, where special tokens are used to encode operation spans.

\subsection{Jointly-Trained Model (\textsc{Multi-task})}
\label{sec:multitask}
To explore the benefits of jointly modeling mentions and relations, we experiment with a graph-based multi-task framework based on \textsc{DyGIE++} model~\cite{wadden-etal-2019-entity}.  Candidate mention spans are encoded using SciBERT, and a graph is constructed based on predicted X-WLP relations and argument roles. A message-passing neural network  is then used to predict mention spans while propagating information about related spans in the graph~\citep{dai2016discriminative,gilmer2017neural,jin2018junction}.

This approach requires computing hidden state representations for all $O(n^4)$ pairs of spans in an input text, which for long sequences, will exhaust GPU memory. While \citet{wadden-etal-2019-entity} considered primarily within-sentence relations, our model must consider relations across the entire protocol, which makes this a problem of practical concern. To address this, we encode a sliding window of $w$ adjacent sentences when the full protocol does not fit into memory, allowing smaller windows for the start and end of the protocol, and concatenate sentences within each window as inputs to the model. 
As a result, each sentence is involved in $w$ windows leading to repeated, possibly contradicting predictions for both mentions and relations. To handle this, we  
output predictions agreed upon by at least $k$ windows, where $k$ is a hyperparameter tuned on a development set.

\section{Experiments}
\label{sec:experiments}
In \S\ref{sec:models}, we presented a pipelined approach to PEG prediction based on SciBERT and a message-passing neural network that jointly learns span and relation representations.
Next, we describe the details of our experiments and present empirical results demonstrating that X-WLP supports training models that can predict PEGs from natural language instructions.

\paragraph{Data.}
\corpusname{} is our main dataset including 279 fully annotated protocols. Statistics of X-WLP are presented in Table \ref{tab:stats}. Additionally, we have 344 protocols from the original WLP dataset.  We use this auxiliary data only for training mention taggers in the pipeline model, and use X-WLP for all other tasks.  For argument role labeling and temporal ordering, negative instances are generated by enumerating all possible mention pairs whose types appear at least once in the gold data.  We use 5-fold cross validation; 2 folds (112 protocols) are used for development, and the other 3 folds (167 protocols) are used to report final results.

\paragraph{Model setup.}
The \textsc{Pipeline} framework employs a separate model for each task, by default using the propagated predictions from previous tasks as input.
In addition, we evaluate the model for each task with gold input denoted as \textsc{Pipeline}\texttt{(gold)}. 
Finally, the \textsc{Multi-task} framework learns all tasks together and we decompose its performance into the component subtasks.

\paragraph{Implementation details.}
We use the uncased version of SciBERT\footnote{\url{https://github.com/allenai/scibert}} for all our models due to the importance of in-domain pre-training. The models under the \textsc{Pipeline} system are implemented using Huggingface Transformers \cite{wolf-etal-2020-transformers}, and we use AdamW with the learning rate $2 \times 10^{-5} $ for SciBERT finetuing. For the \textsc{Multi-task} framework, we set the widow size $w$ to 5, the maximum value that enables the model to fit in GPU memory. For all other hyperparameters, we follow the settings of the WLP experiments in~\cite{wadden-etal-2019-entity}.

\makeatletter
\def\thickhline{%
  \noalign{\ifnum0=`}\fi\hrule \@height \thickarrayrulewidth \futurelet
   \reserved@a\@xthickhline}
\def\@xthickhline{\ifx\reserved@a\thickhline
               \vskip\doublerulesep
               \vskip-\thickarrayrulewidth
             \fi
      \ifnum0=`{\fi}}
\makeatother

\newlength{\thickarrayrulewidth}
\setlength{\thickarrayrulewidth}{2\arrayrulewidth}

\subsection{Results}

\begin{table}[t!]
\small
\begin{center}
\begin{tabular}{clc}
\thickhline Data Split & System & $F_1$  \\ \hline
\multirow{3}{*}{original} & \citet{kulkarni-etal-2018-annotated} &  78.0 \\
 & \citet{wadden-etal-2019-entity} & \textbf{79.7} \\ 
& \textsc{Pipeline} & 78.3  \\ \hline
X-WLP-eval & \textsc{Pipeline} & 74.7  \\
\thickhline
\end{tabular}
\end{center}
\caption{\label{tab:mi_results} Mention identification test set F$_1$ scores for models on the WLP dataset. Top: WLP dataset with the original train/dev/test split. Bottom: excluding X-WLP protocols from the WLP training data, and using them for evaluation.}
\end{table}
\begin{table}[t!]
\small
\begin{center}
\begin{tabular}{lccc}
\thickhline System & P & R & $F_1$  \\ \hline
\textsc{Multi-task} & 76.0 & 69.0 & 72.3 \\ 
\textsc{Pipeline} & 71.8 & 76.3 & 74.0  \\
{ $\bullet$ w/  gold mentions} & {79.0} & {80.2} & {79.6} \\ 
\thickhline
\end{tabular}
\end{center}
\caption{\label{tab:oc_results} Predicate grounding test set results.}
\end{table}
\begin{table}[t!]
\small
\begin{center}
\setlength\tabcolsep{3pt}
\resizebox{\columnwidth}{!}{
\begin{tabular}{lccc}
\thickhline Task & \textsc{Multi-task} & \textsc{Pipeline} & \# gold \\ \hline

{Core} &  &  & \\ 
{ $\bullet$ All roles} & \textbf{57.9} &  53.7 & 2839 \\ 
{ $\bullet$ All roles (gold mentions)} & - & 76.5 &  2839\\ 
{ $\bullet$ ARG0} & \textbf{61.0} & 57.1 & 2313 \\ 
{ $\bullet$ ARG1} & \textbf{36.1} & 32.9 & 412 \\
{ $\bullet$ ARG2} & \textbf{69.7} & 61.4 & 114 \\ \hline

{Non-Core} &  & & \\
{ $\bullet$ All roles} & \textbf{55.7} & 48.8 & 4826 \\
{ $\bullet$ All roles (gold mentions)} & - & 78.1 & 4826 \\ 
{ $\bullet$ site} & \textbf{58.7} & 55.4 & 962 \\ 
{ $\bullet$ setting} & \textbf{77.4} & 74.7 & 974 \\ 
{ $\bullet$ usage} & \textbf{35.6} & 33.0 & 296 \\
{ $\bullet$ co-ref} & \textbf{39.8} & 36.7 & 1014 \\ 
{ $\bullet$ measure} & \textbf{63.3} & 56.6 & 804 \\
{ $\bullet$ modifier} & \textbf{51.0} & 41.8 & 519 \\ 
{ $\bullet$ located-at} & 9.7 & \textbf{13.3} & 179 \\ 
{ $\bullet$ part-of} & 0.5 & \textbf{10.8} & 78 \\ \hline

{Temporal Ordering} & \textbf{61.8} & 57.3 & 2176 \\
{Temp. Ord. (gold mentions)} & - & 76.3 & 2176 \\

\thickhline
\end{tabular}}
\end{center}
\caption{\label{tab:arl_results} Operation argument role labeling (core and non-core roles, decomposed by relation) and temporal ordering test set F$_1$ performance.}
\end{table}
\begin{table}[t!]
\small
\begin{center}
\begin{tabular}{lccc}
\thickhline Split & \textsc{Multi-task} & \textsc{Pipeline} & \# gold  \\ \hline

Intra-sentence & \textbf{63.4} & 58.2 & 2160 \\ 
Inter-sentence & \textbf{32.5} & 39.1 & 679 \\

\thickhline
\end{tabular}
\end{center}
\caption{\label{tab:arl_dist_results} Operation argument role labeling (core roles) test set F$_1$, decomposed based on whether the operation and the argument are triggered within the same sentence (intra-sentence) versus different sentences (inter-sentence).}
\end{table}

The results of the two models on the different subtasks are presented in Tables~\ref{tab:mi_results}-~\ref{tab:arl_dist_results}. We identify three main observations based on these results.

First, \textsc{Pipeline} outperforms \textsc{Multi-task} on the operation classification task in Table \ref{tab:oc_results}, as it uses all protocols from  WLP as additional training data to improve mention tagging.

Second, \textsc{Multi-task} performs better than the \textsc{Pipeline} approach on most relation classification tasks in Table \ref{tab:arl_results}, but is worse than \textsc{Pipeline} when \textsc{Pipeline} uses gold mentions, demonstrating that jointly modeling mentions and relations helps in mitigating error propagation.

Third, \emph{cross-sentence} relations are challenging for both models, as shown in Table~\ref{tab:arl_dist_results}. This explains the low performance of \texttt{co-ref}, which is comprised of 92.4\% cross-sentence relations.

In addition, there are a couple of interesting points to note. In Table~\ref{tab:mi_results}, the performance of \textsc{Pipeline} on the X-WLP subset is lower than its performance on the WLP test set, likely because there are fewer protocols in the training set. For the relation-decomposed performance in Table~\ref{tab:arl_results}, we can see that some of the relations like \peglabel{ARG2} can be correctly predicted by \textsc{Multi-task} using only a few gold labels while some more widely used relations are harder to learn, such as \peglabel{ARG0} and \peglabel{site}; indeed, \peglabel{ARG2} is only used in the \pegpred{spin} operation (see Table \ref{tab:core-rels}), while the other roles participate in more diverse contexts.

\section{Related Work}
\label{sec:related-work}
Natural Language Processing (NLP) for scientific procedural text is a rapidly growing field. To-date, most approaches have focused on text-mining applications~\citep{Isayev2019} and typically annotate only shallow, sentence-level semantic structures (e.g., Fig. \ref{fig:example}, top). Examples include 
WLP~\citep{kulkarni-etal-2018-annotated} and materials science procedures~\citep{mysore2019materials,kuniyoshi-etal-2020-annotating}. Recent interest in automation of lab procedures has also led to sentence-level annotation of procedural texts with action sequences designed to facilitate execution~\citep{Vaucher2020}. 

However, as noted in recent concurrent work~\citep{Mehr101}, neither sentence-level semantic structures nor action sequences are sufficient for the goal of converting text to a machine-executable synthesis procedure; for this purpose, a more structured, process-level semantic representation is required. In particular, executable representations require a structured declaration of the locations and states of the different materials throughout a process, details not represented by sentence-level annotations. Our simulator can naturally represent such information by maintaining a stateful model of the process. Simulation fidelity can be controlled by implementing the execution semantics of operations to the level of detail required. 

\citet{Mehr101} have similarly proposed a process-level executable representation, but use an NLP pipeline consisting primarily of rules and simple pattern matching, relying on a human-in-the-loop for corrections; linking our approach with their framework is a promising future direction.

Structurally, PEGs are similar to abstract meaning representation~(AMR; \citealt{Banarescu2013AbstractMR}), allowing us to use agreement and performance metrics developed for AMR.
In contrast with the sentence-level AMR, a major challenge in this work is annotating and predicting procedure-level representations.\footnote{In addition, in contrast with AMR, PEG nodes are directly mapped to the trigger spans in the document.}

Another line of research focuses on procedural text understanding for more general domains: simple scientific processes~\citep{dalvi-etal-2018-tracking}, open domain procedural texts~\citep{tandon-etal-2020-dataset}, and cooking recipes~\citep{kiddon-etal-2015-mise,bosselut2018simulating}. These works represent process-level information and entity state changes, but typically feature shorter processes, simpler language and an open ontology, compared with our domain-specific terminology and grounded ontology.

Our framework also provides a link to text-based game approaches to procedural text understanding. \citet{tamari-etal-2019-playing} modelled scientific procedures with text-based games but used only synthetic data. Our simulator enables leveraging recent advances on text-based games agents (e.g.,~\citep{adhikari2020learning}) towards \emph{natural} language understanding.

\section{Conclusion}
\label{sec:discussion}
We developed a novel meaning representation and simulation-based annotation interface, enabling the collection of process-level annotations of experimental procedures, as well as two parsers (pipeline and joint modelling) trained on this data.
Our dataset and experiments present several directions for future work, including the modelling of challenging long range dependencies, application of text-based games for procedural text understanding, and extending simulation-based annotation to new domains.

\section*{Acknowledgments}
We would like to thank Peter Clark, Noah Smith, Yoav Goldberg, Dafna Shahaf, and Reut Tsarfaty for many fruitful discussions and helpful comments, as well as the X-WLP annotators: Pranay Methuku, Rider Osentoski, Noah Zhang and Michael Zhan.  This work was partially supported by an Allen Institute for AI Research Gift to Gabriel Stanovsky. This material is based upon work supported by the NSF (IIS-1845670) and the Defense Advanced Research Projects Agency (DARPA) under Contract No. HR001119C0108.  The views, opinions, and/or findings expressed are those of the author(s) and should not be interpreted as representing the official views or policies of the Department of Defense or the U.S. Government.

\bibliography{bib}
\bibliographystyle{acl_natbib}

\newpage
\appendix

\section{Annotation Schema}
\label{sec:ontology}
In the following subsections, we provide further details of the annotation schema used. Section \S\ref{ssec:ontology-construction} describes how the ontology was constructed based on \ap{}, and \S\ref{ssec:ontology-coverage} provides details on ontology coverage for the \corpusname{} protocols which were chosen for annotation. Section \S\ref{ssec:roles-ontology} details the rules defining valid PEG edges, or what relations can hold between various entity types.
The annotation guidelines given to annotators are available on the project web page.

\subsection{Ontology Construction}
\label{ssec:ontology-construction}
Operation nodes correspond to ``action'' entities in WLP. In \corpusname{}, to facilitate conversion to executable instructions, we further add a fine-grain operation type; for each operation, annotators were required to select the closest operation type, or a \pegpred{general} type if none applied.

To define our operation type ontology, we consulted the \ap{}~\citep{Miles2018} open source standard used for executable biology lab protocols. \ap{} defines 35 different operation types,\footnote{Based on \url{https://github.com/autoprotocol/autoprotocol-python/blob/master/autoprotocol/instruction.py} as of January 2021.} from which we grouped relevant types into higher level clusters; \corpusname{} operation types are broadly aligned with \ap{} operation types, but are more general in scope, to not limit applicability to any one platform. For example, we use a more general \pegpred{measure} operation type rather than the specific types of measurement operations in \ap{} (\pegpred{spectrophotometry}, \pegpred{measure-volume}, etc.). 

\begin{table}[b!]
\small
\setlength{\tabcolsep}{3pt}
\begin{tabular}{@{}ll@{}}
\toprule
X-WLP Operation       & Autoprotocol Instructions                                                                                                                                                                                                                     \\ \midrule
Spin                  & Spin                                                                                                                                                                                                                                          \\
Convert               & N/A                                                                                                                                                                                                                                           \\
Seal                  & Seal, Cover                                                                                                                                                                                                                                   \\
Create                & Oligosynthesize, Provision                                                                                                                                                                                                                    \\
General               & N/A                                                                                                                                                                                                                                           \\
Destroy               & N/A                                                                                                                                                                                                                                           \\
Measure               & \begin{tabular}[c]{@{}l@{}}Absorbance, Fluorescence,\\ Luminescence, IlluminaSeq,\\ SangerSeq, MeasureConcentration,\\ MeasureMass, MeasureVolume,\\ CountCells, Spectrophotometry,\\ FlowCytometry, FlowAnalyze,\\ ImagePlate\end{tabular} \\
Mix                   & Agitate                                                                                                                                                                                                                                       \\
Remove                & Unseal, Uncover                                                                                                                                                                                                                               \\
Temperature Treatment & \begin{tabular}[c]{@{}l@{}}Thermocycle, Incubate,\\ FlashFreeze \end{tabular}                                                                                                                                                                                                            \\
Transfer              & \begin{tabular}[c]{@{}l@{}}AcousticTransfer,\\ MagneticTransfer, \\ Dispense, Provision,\\ LiquidHandle, Autopick\end{tabular}                                                                                                                 \\
Wash                  & N/A                                                                                                                                                                                                                                           \\
Time                  & N/A                                                                                                                                                                                                                                           \\ \bottomrule
\end{tabular}
\caption{\label{tab:ap} Mapping between \corpusname{} operation types and corresponding \ap{} instructions (if any exist). \ap{} operations tend to be more specific as they are intended for machine execution. \corpusname{} protocols are written for humans, so operation types are defined at a higher level of abstraction. }
\end{table}

Table \ref{tab:ap} maps between \corpusname{} operation types and their equivalents in \ap{}, if one exists. The \corpusname{} operation types do not perfectly overlap with \ap{} as the former is written for humans, while the latter is designed for the more constrained domain of robot execution. Accordingly, some operations not currently supported in \ap{} were added, like \pegpred{wash}. See Table \ref{tab:ops} for example mention spans for each \corpusname{} operation type.

The set of supported operations was chosen to maximize coverage over the types of operations found in the sentence-level annotations of WLP (see \S\ref{ssec:ontology-coverage} below for details).

\subsection{Ontology Coverage}
\label{ssec:ontology-coverage}

\begin{figure}
\centering
\includegraphics[width=\columnwidth]{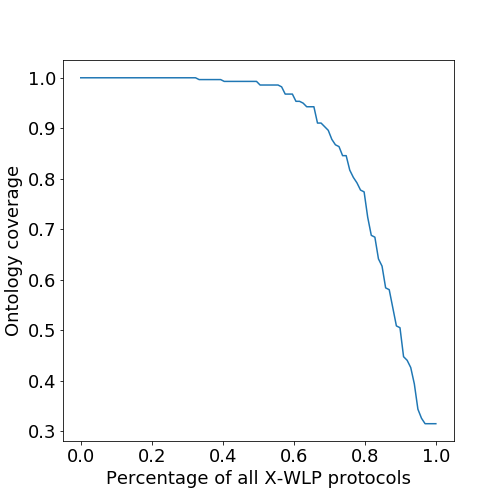}

\caption{\label{fig:ontology-coverage} 
Plot displaying for each coverage percentile ($y$-axis), the percentage ($x$-axis) of \corpusname{} protocols with at least $y$ percent known (ontology-covered) operations.
}
\end{figure}

To identify candidate protocols for annotation which were well covered by the ontology, we created a mapping between ontology instruction types and the 100 most frequent text-spans of WLP action entities (constituting 74\% of all action spans in WLP). WLP action text spans that didn't correspond to any ontology instruction were mapped to a \pegpred{general} label; action text spans that could be mapped to the ontology we call ontology-covered actions. For annotation in X-WLP, we then selected WLP protocols estimated to have a high percentage of ontology-covered actions (based on the mapping above). This simple method was found to be effective in practice, as measured by the actual ontology coverage of X-WLP annotations, summarized in Fig. \ref{fig:ontology-coverage}.

For each annotated protocol, we calculated the percentage of known (not \pegpred{general}) operations. Fig. \ref{fig:ontology-coverage} plots, for each coverage percentile ($y$-axis), the percentage ($x$-axis) of \corpusname{} protocols with at least $y$ percent known operations. From the plot we can see for example that half of the protocols in \corpusname{} have >90\% ontology coverage, and 90\% of the protocols have >70\% ontology coverage.

\subsection{Syntax governing PEG edges}
\label{ssec:roles-ontology}

Formally, edges are represented by triplets of the form $\left(s,r,t\right)$ where $s$ and $t$ are argument nodes and $r$ is a core or non-core role. Dependent on a particular role $r$, certain restrictions may apply to the fine-grained type of $s$ and $t$, as described below.

\subsubsection{Core Roles}
\label{sssec:core-rels}
Core roles, displayed in Table \ref{tab:core-rels}, represent operation specific roles, for example \peglabel{ARG1} for the \pegpred{seal} operation is a \pegarg{seal} entity representing the seal of the \peglabel{ARG0} argument. For core roles, the following restrictions hold:
\begin{compactitem}
    \item  Source nodes $s$ are restricted to any of the \emph{object} types $s \in \left\{ \text{\pegarg{reagent}}, \text{\pegarg{device}}, \text{\pegarg{seal}}, \text{\pegarg{location}}\right\} $ representing physical objects. The only exception to this rule is that \peglabel{ARG1} for the \pegpred{seal} operation must be a \pegarg{seal} entity.
    \item Target node $t$ is a predicate of one of the types in Table \ref{tab:ops}.
    \item $r$ is a core argument relation, $r\in\left\{ \text{ARG0}, \text{ARG1}, \text{ARG2}\right\}$ or ARG* for short.
    \item Certain roles may be required for a valid predicate $t$, for example the \pegpred{transfer} operation requires at minimum both source and target arguments to be specified by the ARG0 and \peglabel{site} roles, respectively.
\end{compactitem}

\subsubsection{Non-core Roles}
\label{sssec:non-core-rels}
\begin{table}[]
\small
\center
\setlength{\tabcolsep}{3pt}
\begin{tabular}{@{}lll@{}}
\toprule
Role    & Source Types   & Target Types                                                              \\ \midrule
co-ref      & Object         & Object                                                                    \\
measure  & Measurement    & Object                                                                    \\
setting  & Setting        & Object                                                                    \\
modifier & Modifier       & \begin{tabular}[c]{@{}l@{}}Object, Operation, \\ Measurement\end{tabular} \\
usage    & Method, Object & Operation                                                                 \\
located-at  & Object         & Object                                                                    \\
part-of     & Object         & Object                                                                    \\ \bottomrule
\end{tabular}
\caption{\label{tab:non-core-rels} Details of non-core roles and restrictions on source and target node types. Object is short for the set of entity types representing physical objects: $\left\{ \text{\pegarg{reagent}}, \text{\pegarg{device}}, \text{\pegarg{seal}}, \text{\pegarg{location}}\right\} $.}
\end{table}
Non-core roles (e.g., \peglabel{setting}, \peglabel{site}, or \peglabel{co-ref}) indicate predicate-agnostic labels. For example, the \pegarg{site} argument always marks the location in which a predicate is taking place. Non-core roles are displayed in Table \ref{tab:non-core-rels}, and role-specific restrictions on $s$ and $t$ are listed under ``Source Types'' and ``Target Types'', respectively.

\end{document}